\pdfoutput=1

\documentclass[11pt]{article}

\usepackage{acl}

\usepackage{times}
\usepackage{latexsym}
\usepackage[normalem]{ulem}
\usepackage{soul}

\usepackage[L7x,T1]{fontenc}
\usepackage[lithuanian,main=english]{babel}

\usepackage[utf8]{inputenc}

\usepackage{microtype}

\usepackage{inconsolata}

\usepackage{amsmath,graphicx}
\usepackage{booktabs}
\usepackage{multirow}
\usepackage{subcaption}
\usepackage{pifont}
\usepackage{inconsolata}

\newcommand{\xmark}{\ding{55}}%
\newcommand{\newpara}[1]{\vspace{1pt}\noindent\textbf{#1}}

\title{UniverSLU: Universal Spoken Language Understanding for Diverse Tasks with Natural Language Instructions}

\author{Siddhant Arora$^{1}$, Hayato Futami$^{2}$, Jee-weon Jung$^{1}$, Yifan Peng$^{1}$, Roshan Sharma$^{1}$${}^*$,\\ {\bf Yosuke Kashiwagi$^{2}$, Emiru Tsunoo$^{2}$, Karen Livescu$^{3}$, Shinji Watanabe$^{1}$}\\
$^{1}$ Carnegie Mellon University, USA\\ 
$^{2}$ Sony Group Corporation, Japan\\
$^{3}$ Toyota Technological Institute at Chicago\\
  \texttt{\{siddhana\}@cs.cmu.edu} \\
}

\begin{document}
\maketitle
\begin{abstract}
Recent studies leverage large language models with multi-tasking capabilities, using natural language prompts to guide the model's behavior and surpassing performance of task-specific models.
Motivated by this, we ask: {\em can we build a single model that jointly performs various spoken language understanding (SLU) tasks?} We start by adapting a pre-trained automatic speech recognition model to additional tasks using single-token task specifiers. We enhance this approach through {\em instruction tuning}, i.e., finetuning by describing the task using natural language instructions followed by the list of label options. Our approach can generalize to new task descriptions for the seen tasks during inference, thereby enhancing its user-friendliness.
We demonstrate the efficacy of our {\em single} multi-task learning model ``UniverSLU'' for 12 speech classification and sequence generation task types spanning 17 datasets and 9 languages.
On most tasks, UniverSLU achieves competitive performance and often even surpasses task-specific models. 
Additionally, we assess the {\em zero-shot} capabilities, finding that the model generalizes to new datasets and languages for seen task types.

\end{abstract}

\section{Introduction}
Spoken language understanding (SLU) refers to the task of inferring semantic meaning or linguistic structure from spoken utterances. Like prior work~\cite{chang2023speechprompt}, we use the term broadly to refer to understanding content~\cite{google_sc}, speaker characterstics~\cite{voxceleb_data}, paralinguistics~\cite{IEMOCAP} and semantics~\cite{SLURP} directly from speech.\footnote{Additionally, we also include audio understanding tasks. Table~\ref{tab:slu-datasets} in Appendix briefly outlines the SLU tasks addressed in this study.\\${}^*$Author is now at Google.} In this work, we define a task as a specific combination of \emph{task type} and \emph{dataset}. A \emph{task type} refers to the type of SLU operation being carried out, like intent classification (IC) or emotion recognition (ER). For instance, performing the same task type (IC) on the SNIPS~\cite{coucke2018snips} and FSC~\cite{Lugosch_FSC} datasets are different tasks.

SLU tasks are traditionally approached with ``task-specific'' models~\cite{whisper_slu},  i.e.~having a separate model for each task. On the other hand, the use of multi-task learning (MTL) approaches in natural language processing (NLP)~\cite{chen2021multitask,multitask_related1} and speech processing~\cite{zhang2023google,whisper} has garnered significant interest.  Such approaches offer cost-efficiency by eliminating the need for separate models for each task. MTL has also been employed for SLU tasks~\cite{ESPnet-SLU,arora-etal-2022-token,huang-etal-2022-mtl}, primarily aiming to improve the performance of tasks with limited labeled data using an auxiliary automatic speech recognition (ASR) objective.

\begin{figure*}[t]
    \begin{subfigure}[t]{0.5\textwidth}
        \centering
        \includegraphics[width=\linewidth]{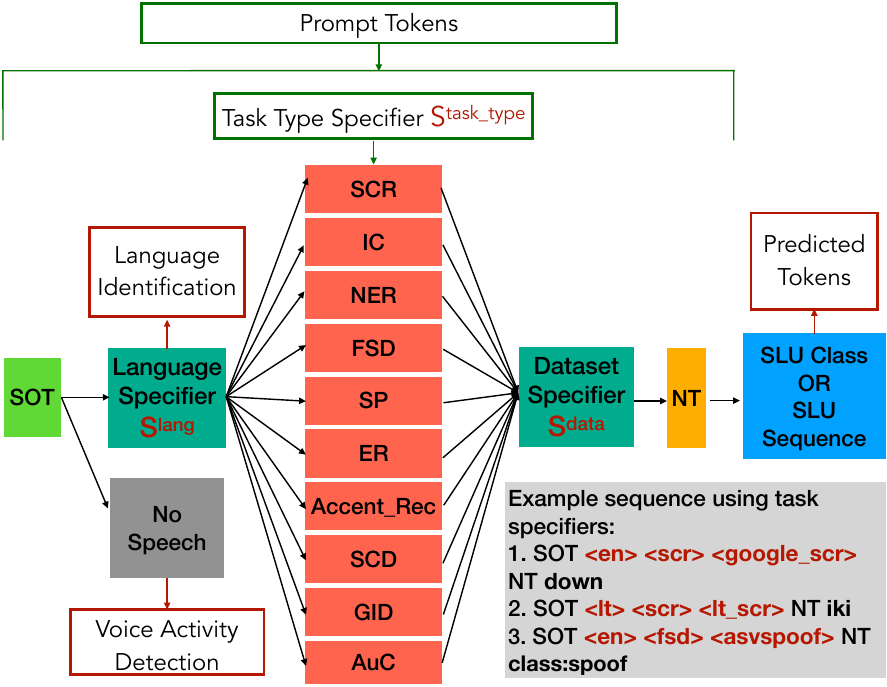}
        \caption{Task Specifiers}
        \label{fig:task specifiers}
    \end{subfigure}
    \begin{subfigure}[t]{0.5\textwidth}
        \centering
        \includegraphics[width=\linewidth]{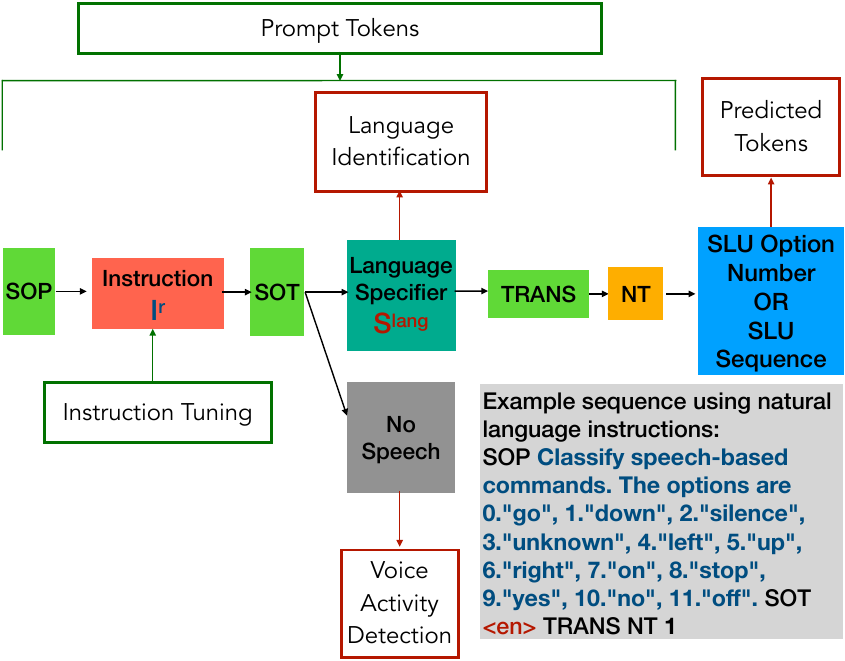}
        \caption{Natural Language Instructions}
        \label{fig:natural instructions}
    \end{subfigure}
    \vspace{-0.2cm}
    \caption{Schematics of our discrete prompt-based MTL formulation. Our architecture comprises an encoder-decoder architecture pre-trained using OpenAI's Whisper model, as detailed in Sec.~\ref{sec: method}. The figure illustrates the sequence of tokens used as prompts and predicted by the decoder during inference. We explore the use of single-token task specifiers ($S^{\text{task\_type}},S^{\text{lang}},S^{\text{data}}$ in Eq.~\ref{eq:task_specifier_formulation}) or natural language phrases ($I^{r}$ in Eq.~\ref{eq:discrete_prompt_viterbi}) to describe the task, as shown in Figures~\ref{fig:task specifiers} and \ref{fig:natural instructions}, respectively. Colored boxes denote a sequence of tokens, while white boxes denote the functionality enabled by the sequence of tokens. SOP, SOT, NT, and TRANS are standard Whisper tokens that specify start-of-prev, start-of-transcript, no-timestamps, and transcribe, respectively.}
    \label{fig:system-overview}
    \vspace{-0.2cm}
\end{figure*}

The rise of large language models (LLMs)~\cite{radford2019language} has sparked interest in MTL, where the selection of appropriate ``prompts''~\cite{prompt1} enables manipulation of the model's behavior to perform specific tasks without additional task-specific parameters. 
LLMs fine-tuned with natural language instructions~\cite{FLAN} describing the task show strong zero-shot performance.
Building upon LLM's success in solving various NLP tasks, there is growing interest in leveraging the ``pre-train, prompt, and predict''~\cite{prompt1} paradigm. 

Prompt-based paradigms have also been explored for speech processing tasks~\cite{WAVPROMPT,peng2023prompting}. SpeechPrompt~\cite{chang2022speechprompt,chang2023speechprompt} employs trainable task-specific prompt vectors with generative spoken language models (GSLMs)~\cite{GSLM}. 
However, this model still lags behind task-specific baselines in most speech classification tasks and particularly struggles on sequence generation tasks. 
Recently, there has been work on instruction tuning for speech processing tasks~\cite{instruction_ASR}. 
However, these approaches mainly concentrate on ASR and related tasks and are often not able to perform diverse SLU tasks.

Motivated by the potential of prompt-based MTL, there has been interest in building ``universal'' ASR models~\cite{zhang2023google,li2021scaling} that can perform ASR across many languages. We extend this work by building a ``universal'' SLU model that can match or even outperform state-of-the-art (SOTA) task-specific SLU models. We leverage the Whisper~\cite{whisper} model, which has been pre-trained on labeled data for ASR and speech translation (ST) tasks using only 2 task specifiers, namely $\langle|$transcribe$|\rangle$ and $\langle|$translate$|\rangle$ respectively. 
We first generalize the Whisper model to a wide range of SLU tasks by utilizing
single token \emph{task specifiers}, as shown in Fig.~\ref{fig:task specifiers}, drawing inspiration from Whisper's pre-training formulation. We show that unlike the most comparable prior multi-task approach, SpeechPrompt v2~\cite{chang2023speechprompt}, our MTL model can outperform or achieve competitive performance to task-specific baselines across many classification and sequence generation tasks. 

However, the use of single-token task specifiers not only constrains our model's ability to handle variations in input format, affecting its user-friendliness, but also prevents it from performing unseen tasks in a zero-shot manner.
Hence, we experiment with utilizing natural language instruction~\cite{FLAN}, such as ``\textcolor{blue}{\textbf{Classify speech-based commands}}'' that describe each task as a prompt (Fig.~\ref{fig:natural instructions}). 
 For classification tasks, we \emph{include a list of options} within the instruction. 
 Our approach generalizes to new paraphrases describing the same task, enhancing the model's user-friendliness. We assess our model's ability to handle unseen tasks, covering new datasets and languages for the known task types, as well as fully unseen task types. This ability is achieved by conducting inference using a task type description accompanied by the list of options specific to the given dataset as the instruction. 
 Our model surpasses random and majority baselines in zero-shot inference on unseen datasets and languages, although it does not match supervised performance.

Our experiments encompass 10 speech classification and 2 sequence generation task types, covering 17 \emph{publicly available} datasets based on 9 languages. Our proposed approach using natural language instruction advances the SOTA on 10 classification tasks and achieves competitive performance to task-specific baselines on sequence generation tasks. We will make our source code and models publicly available as part of the ESPnet-SLU toolkit\footnote{ \url{https://github.com/espnet/espnet}}~\cite{ESPnet-SLU} for reproducibility.
The key contributions of our work are:
\begin{itemize}
    \item We introduce a novel approach for multi-task SLU that leverages human-interpretable instructions combined with a list of option labels as prompts. We demonstrate that our method can generalize to unseen paraphrases.
    \item Our proposed approach can generalize effectively to new datasets and languages for seen task types in a zero-shot manner.
    \item We build a \emph{single} MTL model across 17 SLU datasets that performs competitively or outperforms task-specific models on most tasks.
\end{itemize}

\section{Related Work}
\label{sec:related_work}
\textbf{Multi-task Learning (MTL)} ~\cite{chen2021multitask,Multitask_OG} concurrently trains a model to perform multiple tasks, allowing the model to learn generalizable features instead of task-specific ones. 
MTL has been extensively explored in various NLP settings~\cite{parallel_multitask1,heirarchical_multitask1,modular_multitask1,multitask_adapter1}, ranging from similar task types~\cite{multitask_related2} to the same task type across multiple domains and languages~\cite{multidomain_learning}. MTL can surpass task-specific models in certain scenarios, especially for tasks with limited labeled data~\cite{SpeechStew}.
Similar investigations have been conducted in speech processing, such as adding auxiliary tasks in ASR~\cite{ASR_survey,joint-ctc-att-mtl, Multitask_ASR_survey,toshniwal2017multitask} and for other tasks~\cite{Multitask_ST,multitask_sd,multitask_er,multitask_sr}, primarily using an auxiliary ASR objective. However, these approaches covered very few SLU benchmarks, and we extend this line of research by MTL across 17 SLU datasets.

\textbf{LLMs for speech processing:} 
\label{subsec:LLM}
Recently, there has been significant interest in enhancing LLMs with speech processing and understanding capabilities~\cite{LTU,rubenstein2023audiopalm,fathullah2023prompting,maiti2023voxtlm}. 
These approaches typically involve multi-tasking on a wide range of speech processing tasks, demonstrating robust performance across multiple benchmarks.
LauraGPT~\cite{Laura-GPT} is one such approach that uses Whisper-style task specifiers with Qwen~\cite{bai2023qwen} to handle various speech processing and generation tasks.
Some works have explored preserving LLM capabilities, such as understanding ``instructions'' to perform tasks in a zero-shot manner~\cite{LTU-AS,SALMONN,qwen-audio,fathullah2023towards}. LTU-AS~\cite{LTU-AS} concatenates text instruction with Whisper output and feeds them into Vicuna LLM~\cite{chiang2023vicuna} to perform a range of speech and audio tasks. SALMONN~\cite{SALMONN} employs both speech and audio representation into Vicuna LLM for ``instruction tuning'' across various tasks.

However, while these efforts primarily concentrate on ASR~\cite{yu2023connecting,lakomkin2023end,fathullah2023prompting} and ST~\cite{chen2023salm}, our approach focuses on a diverse array of SLU tasks (see Sec.~\ref{sec:main_results} and \ref{subsec:salmonn_unseen_results}). Furthermore, our novel approach, involving the addition of the list of options in the instruction, not only improves performance but also enhances model generalization to new datasets and languages for the trained task types. Our results in Sec.~\ref{sec:main_results} demonstrate our ability to outperform LLM-based approaches on the subset of SLU tasks they are trained on.
Moreover, we distinguish ourselves as the first to exhibit \emph{instruction tuning capabilities with a decoder of speech foundation model} to the best of our knowledge, that has the important advantage of using much fewer parameters and lower computation.

\section{Problem Formulation}
\label{sec: problem formulation}
Our objective is to build a prompt-based MTL framework capable of performing various speech classification and sequence generation tasks. 
Our model takes a sequence of speech features, denoted $X$, as input for all downstream tasks.
For a specific downstream task $r$, we aim to predict the corresponding $O^r$-length label sequence $Y^r = \{y^r_o \in \mathcal{L}^r | o=1, \dots, O^r\}$, where $\mathcal{L}^r$ represents the label set for that particular task. Table~\ref{tab:slu-datasets} in the Appendix provides a comprehensive description of the diverse SLU tasks covered in this work, including example label sequences $Y^r$ for each task. 
For classification tasks, all label sequences have a fixed length of  $O^r=1$.
To achieve our goal, the prompting paradigm introduces a task-specific prompt $S^r$. Using maximum a posteriori (MAP) decision theory, the MTL model estimates $\hat{Y}^{r}$ as:
\begin{equation}
    \hat{Y}^{r}= \operatorname{argmax} {P(Y^{r}|X, S^{r})} \label{eq:general_formulation}
\end{equation}

\begin{table*}[t]
  \centering
    \resizebox {\linewidth} {!} {
\begin{tabular}{p{20mm}ccp{40mm}p{60mm}}
\toprule
Task Type & Dataset & Language & \multicolumn{2}{c}{\textbf{Generated Prompt}}\\ 
(\textbf{$S^{\text{task\_type}}$}) & (\textbf{$S^{\text{data}}$}) & (\textbf{$S^{\text{lang}}$}) & Task Specifier ($S^{r}$) & Natural Language Instruction ($I^{r}$) \\
\midrule
{\bf Speech Command} & Google SC v1 & En & SOT \textcolor[HTML]{b2182b}{\textbf{$\langle$en$\rangle$ $\langle$scr$\rangle$ $\langle$google\_scr$\rangle$}} NT & SOP \textcolor{blue}{\textbf{Classify speech-based commands. The options are 0."down", $\ldots$}} SOT \textcolor[HTML]{b2182b}{\textbf{$\langle$en$\rangle$}} TRANS NT\\
{\bf Recognition} & Lithuanian SC & Lt & SOT \textcolor[HTML]{b2182b}{\textbf{$\langle$lt$\rangle$ $\langle$scr$\rangle$ $\langle$lt\_scr$\rangle$}} NT & SOP \textcolor{blue}{\textbf{Classify speech-based commands. The options are 0."iki", $\ldots$}} SOT \textcolor[HTML]{b2182b}{\textbf{$\langle$lt$\rangle$}} TRANS NT\\ \midrule
{\bf Fake Speech Detection} & ASVspoof & En & SOT \textcolor[HTML]{b2182b}{\textbf{$\langle$en$\rangle$ $\langle$fsd$\rangle$ $\langle$asvspoof$\rangle$}} NT & SOP \textcolor{blue}{\textbf{Tell apart speech synthesis and voice conversion from authentic speech. The options are 0."bonafide", 1."spoof".}} SOT \textcolor[HTML]{b2182b}{\textbf{$\langle$en$\rangle$}} TRANS NT \\ \midrule
{\bf Accent Classification}& AccentDB & En & SOT \textcolor[HTML]{b2182b}{\textbf{$\langle$en$\rangle$ $\langle$accent\_rec$\rangle$ $\langle$accentdb$\rangle$}} NT & SOP \textcolor{blue}{\textbf{Accent identification in speech. The options are 0."telugu", $\ldots$}} SOT \textcolor[HTML]{b2182b}{\textbf{$\langle$en$\rangle$}} TRANS NT\\ \midrule
{\bf Audio Classification} &  ESC-50 & \xmark& SOT \textcolor[HTML]{b2182b}{\textbf{$\langle$audio$\rangle$ $\langle$auc$\rangle$ $\langle$esc50$\rangle$}} NT & SOP \textcolor{blue}{\textbf{Specify the type of environmental noise in the audio. The options are 0."audio\_class:17", $\ldots$}} SOT \textcolor[HTML]{b2182b}{\textbf{$\langle$audio$\rangle$}} TRANS NT\\  
\bottomrule
\end{tabular}
}
  \caption{Example prompts for different task types, datasets, and languages, with both task specifiers (Eq.~\ref{eq:prompt_task_specifier}) and natural language phrases (Eq.~\ref{eq:prompt_instruction}) as discussed in Sec.~\ref{sec:task_specifier_method} and ~\ref{sec: phrase_method} respectively. \textcolor[HTML]{b2182b}{\textbf{Red bold values}} indicate task specifiers ($S^{\text{task\_type}},S^{\text{lang}},S^{\text{data}}$ in Eq.~\ref{eq:task_specifier_formulation}) and \textcolor{blue}{\textbf{blue bold values}} indicate natural language instruction ($I^{r}$ in Eq.~\ref{eq:discrete_prompt_viterbi}).
  SOP, SOT, NT and TRANS are standard Whisper tokens that specify start-of-prev, start-of-transcript, no-timestamps and transcribe respectively.}
\label{tab:main-sample-utterance}
\end{table*}
 
Prior studies like SpeechPrompt~\cite{chang2022speechprompt,chang2023speechprompt} have modelled the prompting formulation using GSLMs. 
They prepend the embedding (embedding layer $\text{Emb}(\cdot)$) of discrete tokens $U$ with continuous task-specific prompt vectors $P^r$. 
These vectors are learned for each task by maximizing the posterior distribution $P(Y^{r}|\text{Emb}(U), P^{r})$. 

In our approach, instead of relying on a pre-trained model solely for generating discrete tokens for GSLM, we fine-tune an entire pre-trained speech foundation model. Furthermore, we use discrete prompts corresponding to a sequence of tokens.
Starting from a natural extension of the pre-training formulation in speech foundation models like Whisper~\cite{whisper}, we first investigate using single token ``task specifiers'' as prompts.
During pre-training, Whisper utilizes only two task specifiers, namely $\langle$transcribe$\rangle$ and $\langle$translate$\rangle$, to facilitate MTL across ASR and ST tasks, respectively. We experiment with the use of a single token as specifier $S^{\text{task\_type}}$ for each task type, $S^{\text{lang}}$ for each language, and $S^{\text{data}}$ for each dataset as shown in Fig.~\ref{fig:task specifiers} to represent $S^{r}$ (Eq.~\ref{eq:general_formulation}): 
\begin{equation}
    S^{r}=(S^{\text{task\_type}},S^{\text{lang}},S^{\text{data}})
    \label{eq:task_specifier_formulation}
\end{equation}

We then propose to incorporate human-interpretable natural phrases to enhance the attractiveness of our model for human interaction, as shown in the example of Fig.~\ref{fig:natural instructions}. 
We use a probabilistic formulation to provide a new training objective, which handles various natural phrase prompts.
Let $I^{r}$ be the natural language instruction describing downstream task $r$. We can model the posterior distribution in Eq.~\ref{eq:general_formulation} as:
\begin{equation}
    P(Y^{r}|X, S^{r})= \sum_{I^{r}} {P(Y^{r}| X, \text{\sout{$S^{{r}},$}} I^{r})P(I^{r}|\text{\sout{$X,$}}S^{{r}})}\label{eq:discrete_prompt_formulation},
\end{equation}
where we assume conditional independence of $Y^{r}|X, I^{r}$ from $S^{{r}}$ and $I^{r}|S^{{r}}$ from $X$ to simplify Eq.~\ref{eq:discrete_prompt_formulation}.
With the sampling approximation, Eq.~\ref{eq:discrete_prompt_formulation} is rewriten as
\begin{equation}
    P(Y^{r}|X, S^{r})= \sum_{I^{r} \sim P(I^{r}|S^{{r}})} {P(Y^{r}| X, I^{r})}\label{eq:discrete_prompt_viterbi}
\end{equation}
The samples of natural language task descriptions $I^r$ are generated by an LLM that models the distribution $P(I^r|S^r)$, as discussed in Sec.~\ref{sec: phrase_method}.
By using this training scheme, the model can accept various natural phrase prompts provided by the user as a way of describing $S^{{r}}$ (Eq.~\ref{eq:task_specifier_formulation}). 
Unlike prior studies~\cite{chang2022speechprompt,chang2023speechprompt}, our approach allows for potential generalization to various paraphrases $I^{r}$ that specify the same task, thereby enhancing the user-friendliness and adaptability of our model for human interaction.

\section{Prompt-Based Multi-Tasking}
\label{sec: method}
To achieve the formulation described in Eq.~\ref{eq:discrete_prompt_viterbi}, this work proposes a prompt-based MTL training format, as illustrated in Fig.~\ref{fig:natural instructions}.
The input speech $X$ is passed through an encoder-decoder architecture. We prepend instruction $I^{r}$ to the preceding tokens generated by the decoder (shown as ``instruction tuning'' in Fig.~\ref{fig:natural instructions} and described in further detail in Sec.~\ref{sec: phrase_method}). The likelihood of the entire label sequence i.e. $P(Y^{r}|X, I^{r})$ in Eq.~\ref{eq:discrete_prompt_viterbi}, can be computed using task-specific decoder representations $\mathbf{h}_o^r$ as:
\begin{equation}
P(Y^{r}|X, I^{r}) = \prod_{o=1}^{O^r}\text{Softmax}(\text{Out}(\mathbf{h}_o^{r})) \label{final_eq}
\end{equation}
where $\text{Out}(\cdot)$ denotes a linear layer that maps the decoder hidden representation $\mathbf{h}_o^{r}$ to vocabulary $\mathcal{V}$. The vocabulary $\mathcal{V}$ is the union of the label sets $\mathcal{L}^r$ for all the tasks. In this work, both the encoder and decoder are pre-trained using a large speech foundation model~\cite{whisper}.\footnote{To achieve strong performance, we believe that it will be useful to leverage a strong speech foundation model like Whisper that has been trained on large amounts of ASR data. Although in this work, we only experiment with fine-tuning Whisper model, our approach can generalize to incorporating any speech foundation model~\cite{peng2023reproducing}.}

\subsection{Prompting with task specifier}
\label{sec:task_specifier_method}
As discussed in Sec.~\ref{sec: problem formulation}, we also experiment with training our MTL model using task specifiers as shown in Fig.~\ref{fig:task specifiers}. The prompt is constructed based on Eq.~\ref{eq:task_specifier_formulation} as:
\begin{equation}
     \text{Prompt}=\text{SOT}\textcolor[HTML]{b2182b}{\langle\textbf{lang}\rangle \langle\textbf{task}\rangle \langle\textbf{dataset}\rangle} \text{NT} \label{eq:prompt_task_specifier}
\end{equation}
where \textcolor[HTML]{b2182b}{$\langle\textbf{lang}\rangle$} is the language specifier $S^{\text{lang}}$ (Eq.~\ref{eq:task_specifier_formulation}) from a set of language tags, \textcolor[HTML]{b2182b}{$\langle\textbf{task}\rangle$} is the task type specifier $S^{\text{task\_type}}$ (Eq.~\ref{eq:task_specifier_formulation}) from the token set \{scr, ic, ner, $\ldots$\} (denoting speech command recognition, intent classification, named entity recognition, $\ldots$ respectively as further described in Sec.~\ref{subsec:dataset_and_task}), \textcolor[HTML]{b2182b}{$\langle\textbf{dataset}\rangle$} is a dataset specifier $S^{\text{data}}$ in the set \{google\_scr, asvspoof, $\ldots$\} (See Sec.~\ref{subsec:data_appendix}), and SOT and NT represent the standard Whisper tokens specifying start-of-transcript and no-timestamps.

We provide example prompts using task specifiers for some sample tasks in Table~\ref{tab:main-sample-utterance}. Every decoded text sequence must begin with a SOT token. 
The model is then prompted using $S^{\text{lang}}$ as in the pre-training setup~\cite{whisper}, and we also fine-tune the model to predict $S^{\text{lang}}$ for language identification. In addition to the 99 languages already considered during Whisper pre-training, we add another language tag, \textcolor[HTML]{b2182b}{\textbf{$\langle$audio$\rangle$}}, to enable audio-related tasks such as audio classification. If there is no speech in the audio segment and the model is not prompted using the \textcolor[HTML]{b2182b}{\textbf{$\langle$audio$\rangle$}} tag, the model has been pre-trained to predict ``no-speech'' and finish decoding. We fine-tune the model for voice activity detection similar to this pre-training setup.
The next token is the task type specifier $S^{\text{task\_type}}$, and we add a new token to the vocabulary of the pre-trained model for each SLU task type. 
We further add a new token $S^{\text{data}}$ for each dataset since the label set $\mathcal{L}^r$ (see Sec.~\ref{sec: problem formulation}) 
can differ between datasets.  For example, in the case of IC, the label (deactivate\_lights\_none) in~\cite{Lugosch_FSC} indicates the same intent as (switchlightoff) in ~\cite{coucke2018snips}.
Finally, to simplify fine-tuning, we specify that the model should not predict time stamps. The model then uses this prompt to predict the class or label sequence for the given SLU task.
\begin{table*}[t]
  \centering
    \resizebox {\linewidth} {!} {
\begin{tabular}{lccc|ccc|ccc}
\toprule
\multirow{3}{*}{\textbf{Model}}& \multicolumn{3}{c}{\textbf{FSC (IC)}} & \multicolumn{3}{c}{\textbf{IEMOCAP (ER)}} & \multicolumn{3}{c}{\textbf{Google SC v1 (SCR)}}\\
&  &  & Diff & & & Diff & & & Diff\\
& Seen & Unseen & Option & Seen & Unseen & Option & Seen & Unseen & Option\\
& Prompt & Prompt & Order & Prompt & Prompt & Order & Prompt & Prompt & Order\\
 \midrule
SOTA {\footnotesize \em (task-specific models)} & 99.7 & \xmark  & \xmark & 79.2  & \xmark  & \xmark & 98.6 & \xmark  & \xmark \\
 \midrule
Task Specifier (Eq.~\ref{eq:prompt_task_specifier}) & 99.8  & \xmark  & \xmark & 74.7  & \xmark  & \xmark & 99.1 & \xmark  & \xmark \\
Natural Phrase (Eq.~\ref{eq:prompt_instruction}) & 99.7 & 99.7 & 99.7 & 73.9 & 73.5 & 74.2 & 99.1 & 99.1 & 99.1\\
\hphantom{0} Replace ``TRANS'' (Eq.~\ref{eq:ablation_prompt_instruction}) & 99.7 & 99.7 & 99.7 & 72.8 & 73.0 & 73.8 & 99.1 & 99.1 & 99.1\\
\hphantom{0}\hphantom{0}- Options in Instruction & 99.7 &  99.7 & \xmark & 61.5 & 62.0 & \xmark & 99.0 & 99.0 & \xmark \\
\bottomrule
\end{tabular}
}
  \caption{Accuracy (\%)  of the MTL model trained using natural language phrases as prompts on FSC, IEMOCAP and Google Speech Commands dataset.} 
\label{tab:main-paraphrase-results}
\end{table*}
\subsection{Prompting with natural language instructions}
\label{sec: phrase_method}
During training and inference, we query ChatGPT,\footnote{\url{https://chat.openai.com/},\url{https://platform.openai.com/docs/models/gpt-3-5-turbo}} the gpt-3.5-turbo version, to sample natural language paraphrases $I^{r}$ (in Eq.~\ref{eq:discrete_prompt_viterbi}) for the given task specifier $S^{{r}}$. The authors then manually examine these paraphrases and select those believed to be representative of what a user might provide as a task description.\footnote{More details in Sec.~\ref{subsec:chatgpt_prompt}.}
We incorporate instructions $I^{r}$ as previous context text as shown in Fig.~\ref{fig:natural instructions}. 
The prompt for the model trained with natural language instructions is :
\begin{equation}
     \text{Prompt}=\text{SOP}\textcolor{blue}{\langle\textbf{instruction}\rangle}\text{SOT}\textcolor[HTML]{b2182b}{\langle\textbf{lang}\rangle} \text{TRANS NT} \label{eq:prompt_instruction}
\end{equation}
where \textcolor{blue}{$\langle\textbf{instruction}\rangle$} is a natural language instruction $I^{r}$, \textcolor[HTML]{b2182b}{$\langle\textbf{lang}\rangle$} is the language specifier as in Eq.~\ref{eq:prompt_task_specifier}, and the remaining are required standard Whisper tokens (SOP and TRANS specify start-of-prev and transcribe).
For classification tasks, we augment the instruction $I^{r}$ with a list of possible \emph{options} and instruct the model to predict the option number, such as: ``[SOP \textcolor{blue}{\textbf{Classify speech-based commands. The options are 0."go", 1."down", 2."silence", $\ldots$}} SOT \textcolor[HTML]{b2182b}{\textbf{$\langle$en$\rangle$}} TRANS NT 1]''.
As in Sec.~\ref{sec:task_specifier_method}, we fine-tune the model on the pre-training task types, such as language identification and voice activity detection, as in the pre-training setup.

We also provide example prompts using natural instructions for some tasks in Table~\ref{tab:main-sample-utterance}.  
Every decoded text sequence must begin with an SOP token. 
This is followed by natural language instruction followed by a list of possible options. 
We mask out the
training loss over the instruction text and train the
model to predict all other tokens. 
The next token is SOT followed by a language specifier $S^{\text{lang}}$ and TRANS as in its pre-training setup~\cite{whisper}. 
Finally, the model predicts the \emph{option number of SLU class} or label sequence for the given SLU task.

We further perform an ablation study where instead of incorporating instruction as previous context text, we replace the ``TRANS'' token in Eq.~\ref{eq:prompt_instruction} with the instruction phrase, such that
\begin{equation}
     \text{Prompt}=\text{SOT}\textcolor[HTML]{b2182b}{\langle\textbf{lang}\rangle}\textcolor{blue}{\langle\textbf{instruction}\rangle}\text{NT} \label{eq:ablation_prompt_instruction}
\end{equation}
An example model prompt sequence would be ``[SOT \textcolor[HTML]{b2182b}{\textbf{$\langle$en$\rangle$}} \textcolor{blue}{\textbf{Classify speech-based commands. The options are 0."go", 1."down", 2."silence", $\ldots$}} NT]''. We evaluate the efficacy of various ways of incorporating instructions in Sec.~\ref{sec:natural_phrase_result}.

\section{Experiments}
\subsection{Datasets and Tasks}
\label{subsec:dataset_and_task}
We conduct experiments on 18 tasks over 17 publicly available SLU datasets spanning 9 languages and 12 task types, including 10 classification task types (IC, speech command recognition (SCR), language identification (LID), fake speech detection (FSD), emotion recognition (ER), accent classification (AcC), sarcasm detection (SD), gender identification (GID), voice activity detection (VAD), and audio classification (AuC)) as well as 2 sequence generation task types (named entity recognition (NER) and semantic parsing (SP)). We use additional evaluation datasets for testing zero-shot performance in a new dataset and language, as well as for an unseen task type, dialogue act classification (DAC). See Sec.~\ref{subsec:data_appendix} in the Appendix for task type and dataset descriptions. We will make all our data preparation publicly available as part of the ESPnet-SLU toolkit~\cite{ESPnet-SLU} for reproducibility.
\subsection{Baselines}
We compare our approach with SOTA task-specific baselines from prior work~\cite{ESPnet-SLU,googlesc_sota,grabo_sota,msc_sota,fsc_sota,lid_sota,AccentDB,sarcasm_data1,sarcasm_data2,voxceleb_data,vad_sota,auc_sota} by reporting the performance from the original papers (``SOTA'' in Table~\ref{tab:main-classification-results}). We also report the performance achieved by SpeechPrompt v2~\cite{chang2023speechprompt} (``SpeechPrompt v2'' in Table~\ref{tab:main-classification-results}) to quantify the impact of our proposed modifications to prior MTL approaches based on speech prompting, as discussed in Sec.~\ref{sec: problem formulation}.

\subsection{Experimental Setups}
\label{subsec: experiment_setup}
All models are built using the ESPnet-SLU toolkit~\cite{ESPnet-SLU}. We adopt the evaluation metrics used in prior work.

\begin{table*}[t]
  \centering
    \resizebox {\linewidth} {!} {
\begin{tabular}{ccccccc|cccc}
\toprule
\textbf{Task} & \multirow{3}{*}{\textbf{Metric}} & \multirow{3}{*}{\textbf{Dataset}} & \multirow{3}{*}{\textbf{Language}} & \multirow{3}{*}{\textbf{\#Class}} & \multirow{3}{*}{\textbf{SOTA}} & {\textbf{Speech}} & \multicolumn{4}{c}{\textbf{Prompt based MTL}}\\ 
\cmidrule(r){8-11}
\textbf{Type}& & & & & & \textbf{Prompt v2} & \multicolumn{2}{c}{\textbf{UniverSLU-14}} & \multicolumn{2}{c}{\textbf{UniverSLU-17}}\\
& & & & & & & Task & Natural & Task & Natural \\
& & & & & & & Specifier & Phrase & Specifier & Phrase \\
\midrule
\multirow{4}{*}{\bf SCR} & \multirow{4}{*}{Acc $\uparrow$} & Google SC v1 & En & 12 & 98.6 (\citeyear{googlesc_sota}) & \hphantom{0}94.7 & \hphantom{0}99.1 & 99.1 & \hphantom{0}99.1 & \hphantom{0}\textbf{99.2}\\
& & Grabo SC & Du & 36 & 98.9 (\citeyear{grabo_sota})  & \hphantom{0}92.7 & \hphantom{0}\textbf{99.7} & \textbf{99.7} & \hphantom{0}\textbf{99.7} & \hphantom{0}\textbf{99.7}\\
& & Lithuanian SC & Lt & 15 & 91.8 (\citeyear{msc_sota}) & \hphantom{0}95.5 & \textbf{100.0} & 97.7 & \hphantom{0}98.9 & \hphantom{0}98.9\\
& & Arabic SC & Ar & 16 & 98.9 (\citeyear{msc_sota}) & \textbf{100.0} & \hphantom{0}95.9 & 98.9 & \textbf{100.0} & \textbf{100.0}\\
\multirow{3}{*}{\bf IC} &  Acc $\uparrow$ &  Fluent SC & En & 24 & 99.7 (\citeyear{fsc_sota}) & \hphantom{0}98.2 & \hphantom{0}\textbf{99.8} & 99.7 & \hphantom{0}\textbf{99.8} & \hphantom{0}99.7\\
&  F1 $\uparrow$ &  SNIPS & En & \hphantom{0}6 & 96.3 (Tab\ref{tab:main-results}) & \xmark & \xmark & \xmark & \hphantom{0}92.3 & \hphantom{0}63.5\\
&  Acc $\uparrow$ &  SLURP & En & 69 & 89.6 (Tab\ref{tab:main-results}) & \xmark & \xmark & \xmark & \hphantom{0}\textbf{90.3} & \hphantom{0}86.3\\
\multirow{2}{*}{{\bf LID}} & \multirow{2}{*}{Acc $\uparrow$} & \multirow{2}{*}{Voxforge} & En, Es, Fr, & \multirow{2}{*}{\hphantom{0}6} & \multirow{2}{*}{99.8 (\citeyear{lid_sota})} & 
\multirow{2}{*}{\hphantom{0}94.2} & \multirow{2}{*}{\hphantom{0}\textbf{99.9}} & \multirow{2}{*}{\textbf{99.9}} & \multirow{2}{*}{\hphantom{0}\textbf{99.9}} & \multirow{2}{*}{\hphantom{0}\textbf{99.9}}\\
& & &  De, Ru, It & & & \\
{\bf FSD} & EER $\downarrow$ & ASVspoof & En & \hphantom{0}2 & \hphantom{0}2.5 (\citeyear{lid_sota}) & \hphantom{0}13.1 & \hphantom{00}1.0 & \hphantom{0}\textbf{0.8} & \hphantom{00}2.0 & \hphantom{00}0.9\\ 
{\bf ER} & Acc $\uparrow$ & IEMOCAP & En & \hphantom{0}4 & \textbf{79.2} (\citeyear{lid_sota}) & \hphantom{0}50.2 & \hphantom{0}73.4 & 72.3 & \hphantom{0}74.7 & \hphantom{0}73.9\\  
{\bf AcC} & Acc $\uparrow$ & AccentDB & En & \hphantom{0}9 & 99.5 (\citeyear{AccentDB}) & \hphantom{0}87.1 &	\textbf{100.0} & 99.9 & \hphantom{0}99.9 & \textbf{100.0}\\    
\multirow{2}{*}{\bf SD} & \multirow{2}{*}{F1 $\uparrow$} & MUStARD & En & \hphantom{0}2 & 64.6 (\citeyear{sarcasm_data1}) & \hphantom{0}\textbf{78.7} & \hphantom{0}73.2 & 66.7 & \hphantom{0}73.5 & \hphantom{0}72.9\\ 
& & MUStARD++ & En & \hphantom{0}2 & 65.2 (\citeyear{sarcasm_data2}) & \hphantom{0}\textbf{75.2} & \hphantom{0}67.4 & 66.2 & \hphantom{0}73.6 & \hphantom{0}66.3\\
{\bf GID} & F1 $\uparrow$ & VoxCeleb1 & En & \hphantom{0}2 & 98.8 (\citeyear{voxceleb_data}) & \hphantom{0}91.6 &	\hphantom{0}\textbf{99.9} & \textbf{99.9} & \hphantom{0}\textbf{99.9}& \hphantom{0}\textbf{99.9}\\ 
\multirow{2}{*}{{\bf VAD}} & \multirow{2}{*}{Acc $\uparrow$} & Google SC v2 & \multirow{2}{*}{En} & \multirow{2}{*}{\hphantom{0}2} & \multirow{2}{*}{98.8 (\citeyear{vad_sota})} & \multirow{2}{*}{\hphantom{0}98.3} &		\multirow{2}{*}{\hphantom{0}99.0} & \multirow{2}{*}{99.0} & 
\multirow{2}{*}{\hphantom{0}98.8} & \multirow{2}{*}{\hphantom{0}\textbf{99.1}}\\  
& & + Freesound &  &  &  &  &	 &  &  & \\  
{\bf AuC} & Acc $\uparrow$ &  ESC-50 & \xmark & 50 & \textbf{97.0} (\citeyear{auc_sota}) & \hphantom{0}37.5 &			\hphantom{0}39.5 & \hphantom{0}2.0 & \hphantom{0}73.0 & \hphantom{00}2.0\\   
{\bf NER} & SLU F1 $\uparrow$ &  SLURP & En & \xmark & \textbf{79.7} (Tab\ref{tab:main-results}) & \xmark &			\xmark & \xmark & \hphantom{0}79.5 & \hphantom{0}74.8\\   
{\bf SP} & EM $\uparrow$ &  STOP & En & \xmark & \textbf{78.8} (\citeyear{STOP_Track1}) & \xmark &			\xmark & \xmark & \hphantom{0}78.4 & \hphantom{0}75.3\\
\bottomrule
\end{tabular}
}
  \caption{Performance of \textit{UniverSLU} employing both task specifiers (Eq.~\ref{eq:prompt_task_specifier}) and natural language instructions (Eq.~\ref{eq:prompt_instruction}) on both classification and sequence generation tasks, compared to both task-specific SOTA performance (including Whisper task-specific baselines in Tab. 6 in Appendix) and SpeechPrompt v2~\cite{chang2023speechprompt}.} 
\label{tab:main-classification-results}
\end{table*}

As a proof of concept, we initially train an MTL model (refer to Sec.~\ref{sec: phrase_method}) employing natural language instructions $I^{r}$ (in Eq.~\ref{eq:discrete_prompt_viterbi}) as prompts for three tasks using the FSC (IC), IEMOCAP (ER), and Google SC v1 (SCR) datasets. We explore various approaches for incorporating natural language phrases, as discussed in Sec.~\ref{sec: phrase_method}. When the list of options is included in the instruction, we introduce a random shuffling of the order of options during training. Our preliminary investigations (not shown) indicate that this shuffling enhances the model's stability across different option orders in the instruction and improves its generalizability to unseen tasks.

Subsequently, we perform multitasking on the same \emph{14} speech classification datasets as SpeechPrompt v2~\cite{chang2023speechprompt}, utilizing both single-token task specifiers (Eq.~\ref{eq:prompt_task_specifier}) and natural language instructions (Eq.~\ref{eq:prompt_instruction}). We refer to these models as \textit{UniverSLU-14 Task Specifier} and \textit{UniverSLU-14 Natural Phrase}, respectively.
We then expand the training tasks to include sequence generation task types such as NER and SP. This enhanced MTL model, trained on a total of \emph{17} datasets, is denoted \textit{UniverSLU-17}. 
To enhance performance on low-resource and para-linguistic datasets, we upsample (see Sec.~\ref{subsec:appendix_experiment_setup}) the Lithuanian and Arabic speech command datasets, ESC-50, SNIPS, sarcasm detection, and emotion recognition datasets.
All model, training, and inference parameters are selected based on validation performance. Further details are provided in Sec.~\ref{subsec:appendix_experiment_setup}.

\section{Results and Discussion}
\label{sec:results}

\subsection{Using Natural Language Instructions}
\label{sec:natural_phrase_result}

Table~\ref{tab:main-paraphrase-results} shows the performance of our MTL model trained with natural language phrases instead of task specifiers as discrete prompts.
As detailed in Sec.~\ref{sec: phrase_method}, we experiment with various methods of incorporating instructions. We initially report results on seen prompts—natural language instructions encountered during training. In the scenario where the ``TRANS'' token is replaced with natural instructions (Eq.~\ref{eq:ablation_prompt_instruction}), including the list of options in the instruction proves beneficial for performance improvement. Incorporating the natural language instruction as a previous text token (Eq.~\ref{eq:prompt_instruction}) yields the best performance, closely resembling that achieved using task specifiers (Eq.~\ref{eq:prompt_task_specifier}).

To assess the generalizability of our approach, we evaluate the model on natural language phrases not included in training.
We average the results obtained from 5 \emph{unseen natural language phrases}. 
We ensure that these natural phrases are sufficiently different from training prompts using manual inspection and by making sure that the average edit distance of unseen prompt from training prompts, normalized by the length of the unseen prompt, exceeds 0.9 for all datasets.
See the complete list of seen and unseen instructions in Table~\ref{tab:nlp_prompt_detail} in the Appendix.
Our findings suggest that the model generalizes effectively to unseen natural language phrases, exhibiting performance comparable to that with seen prompts. Additionally, we test the model on two different orders of options in the instruction and report the average performance, noting stable results regardless of option ordering.

This investigation underscores the potential of using natural language phrases as prompts.
Results support our decision to depart from continuous prompts used in prior work~\cite{chang2023speechprompt}, opting for natural language instructions to create a more user-friendly model capable of handling diverse paraphrases.

\begin{table*}[t]
  \centering
     
    {
\begin{tabular}{cccccc|cc}
\toprule
 \multirow{2}{*}{\textbf{Task Type}} & \multirow{2}{*}{\textbf{Metric}} & \multirow{2}{*}{\textbf{Dataset}} & \textbf{LauraGPT} & \textbf{LTU-AS} & \textbf{SALMONN} & \multicolumn{2}{c}{\textbf{UniverSLU-17}}\\ 
 & & &  (\citeyear{Laura-GPT}) &  (\citeyear{LTU-AS}) & (\citeyear{SALMONN}) & & \\
& & & Task & Natural & Natural & Task & Natural\\
& & & Specifier & Phrase & Phrase & Specifier & Phrase\\
\midrule
{\bf IC} &  Acc $\uparrow$ &  SLURP & 87.87 & \xmark & \xmark & \textbf{90.3} & 86.3\\
{\bf ER} & Acc $\uparrow$ & IEMOCAP & \xmark & 65.2 & 69.0 & \textbf{74.7} & 73.9\\  
{\bf GID} & F1 $\uparrow$ & VoxCeleb1 & \xmark & 90.8 & \xmark & \textbf{99.9}& \textbf{99.9}\\  
{\bf AuC} & Acc $\uparrow$ &  ESC-50  & \xmark & \textbf{80.8} & \xmark & 73.0 & \hphantom{0}2.0\\   
{\bf NER} & SLU F1 $\uparrow$ &  SLURP & 73.45 & \xmark & \xmark &	\textbf{79.5} & 74.8\\   
\bottomrule
\end{tabular}
}
  \caption{Results comparing \emph{UniverSLU} with recently proposed LLMs for speech processing.} 
\label{tab:main-LLM-results}
\end{table*}
\subsection{More tasks: UniverSLU}
\label{sec:main_results}
We present the performance of our \textit{UniverSLU Task Specifier}, \textit{UniverSLU Natural Phrase}, SpeechPrompt v2, and the SOTA models on these benchmarks in Table~\ref{tab:main-classification-results}.
Our results demonstrate that the \textit{UniverSLU-14 Task Specifier} outperforms prior speech prompting methods on 11 out of 14 tasks and improves the SOTA on 9 out of 14 tasks. 
\textit{UniverSLU-14 Natural Phrase} achieves performance comparable to \textit{UniverSLU-14 Task Specifier} across most tasks, with the added benefit of generalizing to new paraphrases of instructions, as discussed in Sec.~\ref{sec:natural_phrase_result}. Our \textit{UniverSLU-17 Task Specifier} model, trained on a combination of speech classification and sequence generation tasks, demonstrates comparable or superior performance compared to the SOTA on 11 tasks. It also achieves competitive performance compared to task-specific baselines on sequence generation tasks, in contrast to previous approaches~\cite{chang2022speechprompt} that face challenges in handling sequence generation tasks.
We observe that our upsampling approach proves to be highly effective, significantly improving the performance on low-resource datasets such as Arabic speech command recognition. Finally, our \textit{UniverSLU-17 Natural Phrase} model also outperforms the SOTA on 10 tasks and again achieves comparable performance to the \textit{UniverSLU-17 Task Specifier} model on most tasks. These results demonstrate our model's capability to perform very well, exceeding SOTA results even on tasks with limited fine-tuning data, such as AcC, which only involves 20 hours of fine-tuning data. We hypothesize that this strong performance can be attributed to both the model's pre-training as well as our multi-task learning setup.

However, \textit{UniverSLU-14 Natural Phrase} and \textit{UniverSLU-17 Natural Phrase} fail to perform better than the random baseline on the audio classification (AuC) task, presumably because Whisper has not been trained on non-speech tasks. 
Previous work on LTU-AS~\cite{LTU-AS} has shown the capability of Whisper to follow instructions for performing audio-related tasks. However, it's important to note that LTU-AS was fine-tuned on a substantial dataset comprising 845,000 audio clips, while our fine-tuning dataset, ESC-50, is much smaller, consisting of only 1600 training clips.
Based on these results, we hypothesize that pre-training plays a crucial role in effective instruction tuning, particularly when dealing with a small fine-tuning dataset.
\begin{table}[t]
  \centering
    \resizebox {\linewidth} {!} {
\begin{tabular}{lccc}
\toprule
\multirow{5}{*}{\textbf{Model}} & \textbf{Dataset} & \textbf{Language} & \textbf{Task Type}\\
\cmidrule{2-4}
& \textbf{Snips} & \textbf{KSU} & \textbf{Daily}\\
& & \textbf{Emotions} & \textbf{Talks}\\
& \textbf{IC} & \textbf{ER} & \textbf{DAC}\\
 & F1 $\uparrow$ & Acc $\uparrow$ & Acc $\uparrow$\\ 
 \midrule
SOTA & \multirow{2}{*}{96.3} & \multirow{2}{*}{86.8} & \multirow{2}{*}{61.1}\\
{\em(supervised)} & & & \\
 \midrule
 Random & 19.7 & 25.0 & 25.0\\
Majority & \hphantom{0}6.4 & 28.5 & 45.5\\
 \midrule
UniverSLU-14 & \multirow{2}{*}{44.6} & \multirow{2}{*}{38.8} & \multirow{2}{*}{17.6}\\
 (Natural Phrase) & & & \\
UniverSLU-17 & \multirow{2}{*}{\xmark} & \multirow{2}{*}{41.7} & \multirow{2}{*}{19.3}\\
 (Natural Phrase) & & & \\
\bottomrule
\end{tabular}
}
  \caption{Performance of MTL models trained using natural language phrases as prompts on an unseen dataset, language and task type in a zero-shot manner. We include random and majority baselines for comparison and report the performance of supervised models (Table ~\ref{tab:main-results} for SNIPS, \citeyear{KSU_emotion} for KSU Emotions, \citeyear{dynamic_superb} for DAC).} 
\label{tab:main-zero-shot-paraphrase-results}
\end{table}

\newpara{Comparison to recent LLM-based approaches:} As outlined in Sec.~\ref{sec:related_work}, several recent works aim to enhance LLMs with speech processing capabilities. Although the majority of these works do not specifically target SLU tasks, some~\cite{LTU-AS,SALMONN,Laura-GPT} are fine-tuned on a subset of the SLU tasks considered in this work, and we show their comparison with UniverSLU in Table~\ref{tab:main-LLM-results}.\footnote{
SALMONN also reports performance on SLURP; however, it focuses on the slot filling task, which involves predicting only entity mentions. In contrast, the NER results reported in this paper aim to predict both the entity label and its corresponding mention.} 
UniverSLU-17 outperforms all of these approaches on these tasks, except for AuC where LTU-AS excels due to fine-tuning on a larger audio dataset. 
We also use SALMONN's publicly available checkpoint~\cite{SALMONN} and report its zero-shot performance on other UniverSLU tasks in Sec.~\ref{subsec:salmonn_unseen_results}, where it is observed to lag behind UniverSLU models.
In conclusion, our MTL approach exhibits substantial advancements over task-specific models as well as previous MTL approaches, including models that deploy LLMs. 

\subsection{Zero Shot Performance}
\label{sec:zero_shot}
To assess the generalization capabilities of our multi-task model, we initially evaluate its performance on unseen tasks that assess its ability to handle seen task types on new datasets and languages. For this analysis, we employ the SNIPS and KSU\_Emotions\footnote{We consider only 4 emotions for evaluation of all our models as discussed in Sec.~\ref{subsec:data_appendix}.} (Arabic Emotion Recognition) datasets. During inference, similarly to the training process, we include the possible label options in the instruction prompts. We also show the performance of random and majority baselines, obtained by randomly (uniformly) selecting answers and selecting the most common answer, respectively.
Our findings in Table~\ref{tab:main-zero-shot-paraphrase-results} indicate that \textit{UniverSLU-14 Natural Phrase} consistently outperforms random and majority baselines on these tasks. However, a notable performance gap exists when compared to supervised topline models, suggesting substantial room for improvement.

Furthermore, we explore the model's capability to perform a completely new task type—Dialogue Act Classification (DAC)—that was not encountered during training. We observe in Table~\ref{tab:main-zero-shot-paraphrase-results} that Univer-SLU models currently fail to generalize to this unseen task type. 

We also calculate the \emph{following rate} proposed in~\cite{SALMONN}, which measures the percentage of times the instruction-tuned model successfully follows the instructions. We observe that it is 100\% for \textit{UniverSLU} on all tasks, indicating that it always predicts one of the label options. 
This preliminary investigation provides insight into the model's generalization ability to unseen tasks in a zero-shot manner. 

\section{Conclusion}
\label{sec: conclusion}
We introduce an ``instruction tuning'' approach that utilizes human-interpretable natural language instructions, combined with a list of option labels as prompts, to fine-tune a speech foundation model on SLU benchmarks. Our experiments demonstrate the model's ability to understand diverse expressions as instructions during inference. Leveraging this approach, we train a single MTL model named \textit{UniverSLU}, capable of performing 12 SLU task types, encompassing both classification and sequence generation tasks. UniverSLU not only provides cost efficiency but also achieves competitive performance compared to task-specific baselines, outperforming SOTA models on several SLU benchmarks. 
Additionally, we find that the model has some generalization ability on new datasets and languages for seen task types. 

While our current approach faces challenges in generalizing to new task types, it demonstrates strong generalization to new paraphrases describing known tasks (Sec.~\ref{sec:natural_phrase_result}) and achieves non-trivial performance on new datasets and languages with entirely different label sets in a zero-shot scenario. Thus, we believe our approach possesses some instruction-following capabilities.
Prior work, exemplified by instruction tuning models like FLAN~\cite{FLAN}, has shown that generalization improves with an increase in the size of the instruction tuning dataset. We plan to explore this further in future work, along with integrating an LLM-based decoder and implementing few-shot inference techniques.

\section*{Limitations}
Our instruction tuning approach relies on incorporating the list of option labels in the prompt. A potential limitation arises if the option list becomes too long, as it may surpass the token limit of the decoder. We plan to delve deeper into this aspect in future work. Additionally, Table~\ref{tab:main-zero-shot-paraphrase-results} illustrates that our \textit{Univer-SLU} model currently encounters challenges in generalizing to new, unseen task types. 
This may be a limitation of Whisper-style models and could perhaps be addressed with the integration of an LLM-based decoder. It may also be that zero-shot generalization is too challenging for certain unseen tasks, requiring at least a few-shot inference.

\section*{Broader Impact and Ethics}
With our UniverSLU model, we strive to build a single MTL model capable of doing multiple SLU tasks, which saves compute as we do not need separate models for each task. Our approach effectively uses speech foundation models and hence can perform well without requiring large amounts of task-specific labeled data for instruction tuning. Furthermore, we show that our model can perform well in a zero-shot manner for new datasets and languages, which is very useful to ensure speech technologies are available to a wide range of users across various domains. Additionally, we adhere to the ACL Ethics Policy. Our experiments are based on open-source datasets that are widely used in prior work and to our knowledge, there is no violation of privacy in these datasets.

\section*{Acknowledgement}
Experiments of this work used the Bridges2 system at PSC and Delta system at NCSA through allocations CIS210014 and IRI120008P from the Advanced Cyberinfrastructure Coordination Ecosystem: Services \& Support (ACCESS) program, supported by National Science Foundation grants \#2138259,\#tel:2138286, \#tel:2138307, \#tel:2137603, and \#tel:2138296.

\bibliography{anthology,custom}
\newpage
\appendix
\begin{table*}[t]
  \centering
     {
\begin{tabular}{lcccc}
\toprule
\multirow{3}{*}{\textbf{Model}} & \multicolumn{1}{c}{\textbf{SNIPS}} & \multicolumn{2}{c}{\textbf{SLURP}} & \multicolumn{1}{c}{\textbf{STOP}}\\ 
\cmidrule(r){2-2}\cmidrule(r){3-4}\cmidrule(r){5-5}
& \textbf{IC} & \textbf{IC} & \textbf{NER} & \textbf{SP}\\
 & \textbf{F1 $\uparrow$} & \textbf{Acc $\uparrow$} & \textbf{SLU F1 $\uparrow$} & \textbf{EM $\uparrow$}\\ 
 \midrule
\multicolumn{5}{l}{\textbf{Task-specific baselines} {\footnotesize \em (models tuned for each task)}}\\
\hline\hline
ESPnet-SLU & 91.7 &	87.4 & 77.7 & 73.3\\
Whisper task-specific & \textbf{96.3} & 89.6 &	79.7 & \textbf{78.8}\\
\midrule
\multicolumn{5}{l}{\textbf{UniverSLU-3} {\footnotesize \em (single model covers all tasks)}}\\
\hline\hline
 w/ Task Type Specifier $S^{\text{task\_type}}$ & 93.9 & \textbf{90.5} & \textbf{80.5} & 78.4\\
 \hphantom{000} w/ Dataset Specifier $ S^{\text{data}}$ & 92.7 & 90.4 & 80.4 & 78.3 \\
\bottomrule
\end{tabular}
}
  \caption{Results showing performance of prompt based MTL model \textit{UniverSLU-3} trained using task specifiers on selected datasets.} 
\label{tab:main-results}
\end{table*}
\section{Appendix}
\label{sec:appendix}
\subsection{Comparison to Task-Specific Baseline}

Since most of these baselines do not incorporate Whisper, we train ``task-specific'' baselines (see Table~\ref{tab:main-results}) by fine-tuning Whisper on selected datasets, following the approach in \cite{STOP_Track1}, to better understand the efficacy of training multiple tasks simultaneously.
Table~\ref{tab:main-results} presents the results of task-specific baselines and the MTL model (\textit{UniverSLU-3}) trained with task specifiers (Sec.~\ref{sec:task_specifier_method}) on selected speech classification and sequence generation tasks, namely IC, NER, and SP tasks. We start by comparing the task-specific baseline, which involves fine-tuning the Whisper model, with end-to-end SLU models in ESPnet-SLU~\cite{ESPnet-SLU} (rows 1 and 2). Results show that fine-tuning Whisper is beneficial for SLU tasks and yields strong baselines.

MTL framework \textit{UniverSLU-3} that incorporates task specifiers $S^{\text{lang}}, S^{\text{task\_type}}, S^{\text{data}}$ (Eq.~\ref{eq:task_specifier_formulation} and \ref{eq:prompt_task_specifier}) achieves competitive performance to the task-specific baselines\footnote{Except for the SNIPS dataset where the results appear to be slightly unstable due to the limited size of the test set. It includes only 166 utterances.}. 
We additionally investigate training the MTL model using only $S^{\text{task\_type}}$ (w/ Task Type Specifier in Table 3) (e.g., ``\textcolor[HTML]{b2182b}{\textbf{$\langle$en$\rangle$ $\langle$scr$\rangle$}} NT down'') and both the $S^{\text{task\_type}}$ and $S^{\text{data}}$ prompts (w/ Data Specifier in Table 3)(e.g., ``\textcolor[HTML]{b2182b}{\textbf{$\langle$en$\rangle$ $\langle$scr$\rangle$ $\langle$google\_scr$\rangle$}} NT down'' same as Eq.~\ref{eq:prompt_task_specifier}). 
We observe similar performance with and without using the dataset specifier ($S^{\text{data}}$ in Eq.~\ref{eq:prompt_task_specifier}) as an additional prompt. Based on these findings, we conclude that the multi-tasking model achieves comparable performance to task-specific baselines for both classification and sequence generation tasks. We can also significantly reduce the number of trainable parameters since we now have a \emph{single} MTL model instead of multiple models, making it cost-efficient.

\begin{table*}[t]
  \centering
  \resizebox{\linewidth}{!}{
  \begin{tabular}{lllp{40 mm}p{60mm}}
    \toprule
\textbf{Tasks Type} & \textbf{Dataset} & \textbf{Language} & \textbf{Output Type} & \textbf{Example Outputs ($Y^r$)}\\
    \midrule
    \multirow{4}{*}{\bf SCR} & Google SC v1 & En & \multirow{4}{*}{speech command class} & yes, down, no, $\ldots$\\
& Grabo SC & Du &  & turn\_rel throttle="slow" angle="east", turn\_rel throttle="slow" angle="west", turn\_rel throttle="fast" angle="south", $\ldots$\\
& Lithuanian SC & Lt &  & \begin{otherlanguage}{lithuanian}
ačiū, iki, išjunk\end{otherlanguage}, $\ldots$\\
& Arabic SC & Ar & & A, B, C, $\ldots$ \\
\multirow{3}{*}{\bf IC} &  Fluent SC & En &  \multirow{3}{*}{intent class} & increase heat washroom, deactivate lights, deactivate lights bedroom, $\ldots$ \\
&  SNIPS & En & & increase brightness, set light color, set light brightness,  $\ldots$\\
&  SLURP & En & & music likeness, music settings, play music, $\ldots$\\
\multirow{2}{*}{{\bf LID}} & \multirow{2}{*}{Voxforge} & En, Es, Fr, & \multirow{2}{*}{langauge class}& En, Es, Fr,\\
& &  De, Ru, It &  & De, Ru, It\\
{\bf FSD} & ASVspoof & En & synthetic speech class & spoof, bonafide\\ 
{\bf ER} & IEMOCAP & En & emotion class & angry, neutral, sad, happy, other\\  
{\bf AcC} & AccentDB & En & speaker accent class & american, australian, bangla,$\ldots$ \\    
\multirow{2}{*}{\bf SD} & MUStARD & En & \multirow{2}{*}{sarcasm class} & sarcasm, not sarcasm\\ 
& MUStARD++ & En & & sarcasm, not sarcasm\\
{\bf GID} & VoxCeleb1 & En & speaker gender class & male, female \\ 
\multirow{2}{*}{\bf VAD} & Google SC v2 & \multirow{2}{*}{En} & \multirow{2}{*}{voice activity class} & \multirow{2}{*}{speech, no speech}\\ 
& + Freesound & &  & \\ 
{\bf AuC} & ESC-50 & \xmark & audio class & dog, pouring water, breathing, $\ldots$\\   
{\bf NER}&  SLURP & En & sequence of (entity phrase, entity tag) pairs & sl:date FILL tuesday SEP sl:event\_name FILL meeting, $\ldots$\\   
{\bf SP} &  STOP & En & sematic parse sequence & [in:create\_alarm [sl:date\_time\_recurring for six am every day ] ], $\ldots$\\
    \bottomrule
  \end{tabular}
  }
  \caption{Overview of the fine-tuning datasets and task types in the UniverSLU model. Each row indicates a ``separate'' task. Note that a task refers to a specific combination of task type and dataset. For instance, SNIPS and FSC have the same task type but are different tasks due to differences in their label set $\mathcal{L}^r$. We also show the example outputs $Y^r$ (seperated by ``,'') for each task.}
    \label{tab:slu-datasets}
\end{table*}
\subsection{Data and Task Type}
\label{subsec:data_appendix}
\textbf{Intent Classification (IC)} aims to recognize the intent from a user's command in order to take appropriate action. We leverage 2 publicly available corpora, i.e., SNIPS and FSC. The \textbf{SNIPS} corpus~\cite{coucke2018snips} (CC-BY-SA license) consists of 1,660 conversations with in-house voice assistant. For our experiments, we utilize a random split following the approach outlined in~\cite{agrawal2020tie}.
The \textbf{Fluent Speech Commands (FSC)} dataset~\cite{Lugosch_FSC} (Public License) benchmarks IC performance with 30,043 commands spoken to an intelligent home assistant. 

\textbf{Named Entity Recognition (NER)} refers to the task of labeling the spoken tokens in a user's command with associated entities and recognizing mentions of these entities. It is a {\em sequence generation} task. The \textbf{SLURP} corpus~\cite{SLURP} (CC BY-NC 4.0 license) is an open source NER corpus consisting of 57.4 hours of single-turn conversations with a home assistant, making it relevant to commercial SLU applications. To augment our training set, we incorporate synthetic data, following prior work~\cite{ESPnet-SLU}.

\textbf{Semantic Parsing (SP)} is a {\em sequence generation} task that focuses on converting a spoken utterance into a semantic parse sequence, enabling the voice assistant to execute tasks effectively.
The \textbf{STOP} dataset~\cite{stop2022} (STOP Dataset License Agreement\footnote{\url{https://github.com/facebookresearch/spoken_task_oriented_parsing/blob/main/LICENSE}}) is the largest semantic parsing dataset. It comprises over 200,000 audio files from more than 800 speakers across 8 different domains.

\textbf{Speech Command Recognition (SCR)} refers to the task of detecting when a keyword has been spoken in an utterance.
The \textbf{Google Speech Commands} dataset~\cite{google_sc} (CC BY 4.0 license) consists of nearly 100,000 utterances of spoken keywords.
The \textbf{Grabo} dataset~\cite{Grabo_data} (License for use in academic research or commercial product) comprises English and Dutch spoken commands given to a robot for navigation. The dataset includes recordings from 11 speakers issuing 36 different commands, each repeated 15 times. 
The \textbf{Lithuanian SC} dataset~\cite{Lithuanian_SC} (license for use in academic research) consists of recordings of 28 individuals uttering 20 Lithuanian words using a mobile phone.
The \textbf{Arabic SC} database~\cite{arabic_sc} (CC BY 4.0 license) consists of 1,600 speech recordings from 40 speakers in Arabic, covering 6 control words and the digits 0 through 9.

\textbf{Language Identification (LID)} aims to recognize the language of a spoken utterance. \textbf{VoxForge}~\cite{maclean2018voxforge} (GPL license) is an open-source dataset established to collect transcribed multilingual speech using open-source engines. We utilize this dataset to train a language classification system for several European languages, including English, Spanish, French, German, Russian, and Italian. Consistent with previous work ~\cite{chang2023speechprompt}, we use random split with 1200 audio samples in the training, 300 in the validation, and 300 in the test set for each language. 

\textbf{Fake Speech Detection (FSD)}
 aims to detect fake or spoofed speech generated by Text-to-Speech and Voice Conversion systems from real speech in order to prevent adversarial attacks.
\textbf{ASVSpoof}~\cite{ASVspoof_2019} (ODC-By license) is a benchmark designed for building fake speech detection systems. In our experiments, we utilize the Logical Access (LA) portion of the ASVspoof 2019 dataset, which includes both bonafide and spoofed speech. The dataset consists of a total of 121,461 utterances with real speech collected from 107 English speakers. 

\textbf{Emotion Recognition (ER)}
 aims to understand the emotion or sentiment conveyed in an utterance.
\textbf{IEMOCAP}~\cite{IEMOCAP}\footnote{License:\url{https://sail.usc.edu/iemocap/Data_Release_Form_IEMOCAP.pdf}} is a dataset consisting of approximately 12 hours of speech data with four emotion classes: neutral, happy, sad, and angry. We use Sessions 1-4 as training and Session 5 as a test set.

\textbf{Accent Classification (AcC)} aims to classify speech from speakers with different accents within the same language.
\textbf{Accent DB}~\cite{AccentDB} (CC BY-NC 4.0 License) dataset contains approximately 20 hours of English speech data, encompassing 4 non-native accents, 1 metropolitan Indian accent, and 4 native accents.

\textbf{Sarcasm Detection (SD)} aims to identify whether a given utterance is sarcastic or not, based on verbal and non-verbal cues.
The \textbf{MUStARD}~\cite{sarcasm_data1} (MIT License) dataset is a multimodal sarcasm detection dataset consisting of 1,991 utterances compiled from popular TV shows and annotated with sarcasm labels.
The \textbf{MUStARD++}~\cite{sarcasm_data2} (Publicly available) dataset is an enhanced version of MUStARD dataset. It includes original MUStARD dataset and features enhanced annotations. Additionally, the dataset has been doubled in size, comprising 2,695 utterances.

\textbf{Gender Identification (GID)} aims to determine the gender of a speaker based on their speech.
For gender identification, we utilize the \textbf{VoxCeleb1}~\cite{voxceleb_data} (Creative Commons Attribution-ShareAlike 4.0 International License) dataset. It contains approximately 148,000 training and 5,000 test utterances. The gender labels are extracted from the speaker metadata.

\textbf{Voice Activity Detection (VAD)} systems aim to determine whether a given audio contains human speech or background noise.
We leverage the \textbf{Google Speech Commands v2}~\cite{google_sc} (CC BY 4.0 license) dataset as speech data and \textbf{Freesound}~\cite{freesound_data} (GNU Affero General Public License v3.0) dataset as background noise to train these systems.

\textbf{Audio Classification (AuC)} is a multi-class single-label classification task that aims to correctly classify the audio present in the environment.
\textbf{ESC-50}~\cite{freesound_data} (Creative Commons
Attribution-NonCommercial license) is a labeled set of 2000 environment recordings with 50 audio classes.

\textbf{Zero-shot evaluation datasets:} 
To assess the performance of our MTL model in an unseen language, we employ the Arabic Emotion Recognition dataset, KSU\_emotions~\cite{KSU_emotion} (License: LDC User Agreement for Non-Members). We utilize a subset of this dataset that mirrors the emotions present in the IEMOCAP dataset—namely, neutral, happy, sad, and angry—comprising 2301 utterances. Additionally, we extend our evaluation to a new, unseen task type: \textbf{Dialogue Act Classification (DAC)} which aims to identify the function of a given speech utterance in a dialog. For this task, we utilize the Daily Talks~\cite{lee2023dailytalk} (CC-BY-SA 4.0 license) spoken dialogue dataset, annotated with four dialogue act categories—question, inform, directive, and commissive—comprising 4758 utterances, following the setup outlined in~\cite{dynamic_superb}.

Finally, it is important to note that all the datasets are \emph{publicly available}.

\subsection{Experimental Setups}
\label{subsec:appendix_experiment_setup}
Our models are implemented in PyTorch~\cite{pytorch}, and experiments are conducted with ESPnet-SLU toolkit~\cite{ESPnet-SLU}. 

\newpara{Task-specific baselines:} We fine-tune the Whisper medium model with more frequent saving of checkpoints (i.e., after every 1,000 iterations instead of after each epoch) following the approach in ~\cite{STOP_Track1}. We train for 100 epochs with a learning rate of 1e-5 and 1500 warmup steps. Early stopping is performed if the validation accuracy saturates. Additionally, we add special tokens to the vocabulary of the Whisper model for SLU labels, such as slot and intent tags. Similar to prior work~\cite{ESPnet-SLU}, we use an auxiliary ASR objective to train our SLU models when we have access to transcripts.

\newpara{MTL models using task specifiers:} Similar training setup is followed for training our multitasking model with task specifiers $S^{\text{lang}}, S^{\text{task\_type}}, S^{\text{data}}$ (Eq.~\ref{eq:task_specifier_formulation}). 
In addition to SLU tags, we also add $S^{\text{task\_type}}$, $S^{\text{data}}$, and ``$\langle$audio$\rangle$'' tokens to the Whisper vocabulary, which are used as prompts for the model, as discussed in Sec.~\ref{sec:task_specifier_method}. For \textit{UniverSLU-17 Task Specifier} we increase the number of iterations per epoch to 3,000
and train for 100 epochs, since training data becomes larger
due to the combination.
\begin{table}[t]
  \centering
  \resizebox {\linewidth} {!} {
  \begin{tabular}{lr}
    \toprule
    Hyperparameter & Value \\
    \midrule
    Whisper model & medium \\
    Dropout Rate & 0 \\
    LR schedule & inv. sqrt.\\
    Max learning rate & [1e-5, 2e-5, 5e-5] \\
    Warmup steps & [5000, 15000, 25000] \\
    Number of steps & [25, 50, 100] \\
    Adam eps  & 1e-6 \\
    Adam betas  & (0.9, 0.99)\\
    Weight decay & 0.01\\
    \midrule
    Beam Size & [1, 5, 20]\\
    Length Penalty & [0,0.1,0.2]\\
    Maxlen ratio & [1.0,1.2]\\
    \bottomrule
  \end{tabular}
  }
  \caption{Model Training and Inference Search for UniverSLU Models.}
  \label{tab:hp-st-tr}
\end{table}
\begin{table*}[t]
  \centering
    \resizebox {\linewidth} {!} {
\begin{tabular}{cccccc|cc}
\toprule
\multirow{3}{*}{\textbf{Task}} & \multirow{3}{*}{\textbf{Metric}} & \multirow{3}{*}{\textbf{Dataset}} & \multirow{3}{*}{\textbf{Language}} & \multirow{3}{*}{\textbf{\#Class}} & \textbf{SALMONN} & \multicolumn{2}{c}{\textbf{UniverSLU-17}}\\ 
\cmidrule(r){7-8}
& & & & & Natural Phrase & Task Specifier & Natural Phrase \\
\midrule
\multirow{4}{*}{\bf SCR} & \multirow{4}{*}{Acc $\uparrow$} & Google SC v1 & En & 12 & 64.6 & \hphantom{0}99.1 & \hphantom{0}\textbf{99.2}\\
& & Lithuanian SC & Lt & 15 & 18.2 & \hphantom{0}\textbf{98.9} & \hphantom{0}\textbf{98.9}\\
& & Arabic SC & Ar & 16 & \xmark & \textbf{100.0} & \textbf{100.0}\\
{\bf IC} &  F1 $\uparrow$ &  SNIPS & En & \hphantom{0}6 & 10.9 & \hphantom{0}\textbf{92.3} & \hphantom{0}63.5\\
\multirow{2}{*}{{\bf LID}} & \multirow{2}{*}{Acc $\uparrow$} & \multirow{2}{*}{Voxforge} & En, Es, Fr, & \multirow{2}{*}{\hphantom{0}6} & \multirow{2}{*}{30.6} & \multirow{2}{*}{\hphantom{0}\textbf{99.9}} & \multirow{2}{*}{\hphantom{0}\textbf{99.9}}\\
& & &  De, Ru, It & & & \\
{\bf FSD} & EER $\downarrow$ & ASVspoof & En & \hphantom{0}2 & 19.9 & \hphantom{00}2.0 & \hphantom{00}\textbf{0.9}\\
{\bf AcC} & Acc $\uparrow$ & AccentDB & En & \hphantom{0}9 & \hphantom{0}8.7 & \hphantom{0}99.9 & \textbf{100.0}\\    
\multirow{2}{*}{\bf SD} & \multirow{2}{*}{F1 $\uparrow$} & MUStARD & En & \hphantom{0}2 &  37.2 & \hphantom{0}\textbf{73.5} & \hphantom{0}72.9\\ 
& & MUStARD++ & En & \hphantom{0}2 & 36.4 & \hphantom{0}\textbf{73.6} & \hphantom{0}66.3\\
{\bf GID} & F1 $\uparrow$ & VoxCeleb1 & En & \hphantom{0}2 & 98.6 & \hphantom{0}\textbf{99.9}& \hphantom{0}\textbf{99.9}\\ 
\bottomrule
\end{tabular}
}
  \caption{Performance of \emph{SALMONN} using publicly available checkpoint on unseen SLU tasks, that were used for fine-tuning \emph{UniverSLU} models.} 
\label{tab:main-salmonn-results}
\end{table*}

\newpara{MTL models using natural language instruction:} For the proof of concept model, trained using 3 datasets, namely FSC, IEMOCAP and Google SC v1, we get 15 natural language descriptions from ChatGPT and use these descriptions to generate instructions for each spoken utterance. We further perform random shuffling of the order of options and combine each natural language description with 3 different random option orders to generate training utterances. For \textit{UniverSLU-14 Natural Phrase} and \textit{UniverSLU-17 Natural Phrase}, we query ChatGPT to obtain ten unique natural language task descriptions from ChatGPT. Each description is further combined with 2 random option orders to generate training utterances. We require only a reduced number of task descriptions and option orders since the training data is much bigger than the one for the proof of concept model. Since the number of training utterances is much larger than one for \textit{UniverSLU-14 Task Specifier} (nearly 20 times), we increased the number of iterations per epoch to 25,000 and observed the model to converge in 25 epochs. In contrast to \textit{UniverSLU-14 Task Specifier}, we only add ``$\langle$audio$\rangle$'' token to the Whisper vocabulary, which is used to fine-tune on audio classification tasks.

The combined parameter size for all our MTL and task-specific models is approximately 762.3M. We initialize the \textit{UniverSLU-17 Task Specifier} and \textit{UniverSLU-17 Natural Phrase} models with weights obtained from the MTL model \textit{UniverSLU-14 Task Specifier} and \textit{UniverSLU-14 Natural Phrase}, respectively. 
We additionally perform upsampling on low-resource datasets: the Arabic speech commands dataset by a factor of 6, the Lithuanian speech commands dataset, ESC-50, SNIPS, and sarcasm detection datasets by a factor of 3, and the emotion recognition dataset by a factor of 2. We chose these upsampling ratios roughly in inverse proportion to the number of utterances in the datasets. In future work, we will further investigate various data sampling approaches and their impact on model performance.

We apply SpecAugment \cite{specaugment} and use dropout \cite{dropout} and label smoothing \cite{label-smoothing} techniques. The models are trained using 4 NVIDIA A40 (40GB) GPUs. All model, training, and inference parameters are selected based on validation performance. Table ~\ref{tab:hp-st-tr} shows training and inference hyperparameters for our hyperparameter search. Full Details about models, configuration files, and data preparation will be made publicly available as part of the ESPnet-SLU toolkit: \url{https://github.com/espnet/espnet}.

\subsection{Detailed Comparison with LLMs}
\label{subsec:salmonn_unseen_results}

In the evaluation process, we utilized the publicly available SALMONN~\cite{SALMONN} 13B checkpoint and applied it to unseen UniverSLU tasks in a zero shot manner, employing instructions similar to those used in UniverSLU as prompts. The results, however, are presented for a subset of tasks due to various challenges encountered during the evaluation process. Difficulties were faced in obtaining results for sequence prediction tasks~\cite{stop2022}, primarily because the output did not match the expected label format. GPU memory constraints were encountered for classification tasks~\cite{Lugosch_FSC,Grabo_data} with more than 20 labels since the option labels were included in the instruction.
Voice activity detection task with very long audios posed a challenge, leading to out of memory error on a 40 GB GPU.

Table~\ref{tab:main-salmonn-results} shows that SALMONN 13B did not perform well on most tasks. However, it's crucial to recognize that this comparison with Univer-SLU is not entirely fair, as SALMONN is operating in a zero-shot manner. Notably, SALMONN exhibited even poorer performance than random for accent classification, intent classification (SNIPS) and sarcasm detection tasks. It was also unable to perform arabic speech command recognition, never predicting one of the label options. Furthermore, in zero-shot inference on SNIPS, SALMONN performed worse than UniverSLU-14 Natural Phrase (See Tab.~\ref{tab:main-zero-shot-paraphrase-results}).
These results strongly support our assertion that existing LLMs struggle with the challenges posed by diverse SLU tasks. The need for UniverSLU, as demonstrated by its superior performance in handling a wide array of SLU tasks, becomes evident in light of these findings.
\subsection{Generating Instructions from ChatGPT}
\label{subsec:chatgpt_prompt}
In this work, we utilize ChatGPT to generate natural language task descriptions $I^{r}$ for each task type. We list our prompts for obtaining task descriptions for each task type in Table~\ref{tab:chatgpt_prompt_detail}. We further perform manual inspection to ensure that these task descriptions resemble how a user might describe a task. To illustrate, for the voice activity detection (VAD) task type, we manually selected the following example paraphrases such as "Determine the presence of spoken words in the audio.", "Identify if there is speech in the provided audio.", "Verify the existence of speech in the given audio.".
For the VAD task type, we deliberately excluded paraphrases that did not explicitly address the detection of "speech" within the audio. For example, "Appraise if there are audible utterances in the audio" appeared to discuss the possibility of general audio, including background noise like a dog barking, rather than focusing specifically on speech. Similarly, "Assess if there is language spoken in the audio" seemed to lean towards identifying the language being spoken, diverging from the primary objective of VAD. We found these descriptions to deviate somewhat from the essence of the VAD task; hence, we removed them as we believe that the end user would not be using these descriptions to describe that task.

We keep querying the ChatGPT until we have sufficient task descriptions for training and inference of our models. We plan on exploring a more automated and systematic approach that can efficiently scale to select descriptions across a broader range of tasks in future work.
\begin{table*}[!htp]\centering
\scriptsize
\resizebox{\linewidth}{!}{%
\begin{tabular}{p{0.03\linewidth} | p{0.5\linewidth}}
\toprule
\textbf{Task Type} &\textbf{Prompt} \\\midrule
\textbf{SCR} & Different ways of saying ``recognize speech command'' in 15 ways in a natural language without changing meaning\\ 
{\bf IC} &  Different ways of saying ``classify intent of spoken utterance'' in 15 ways in a natural language without changing meaning\\
{\bf LID} & Different ways of saying ``identifying the language in verbal communication.'' in 15 ways in a natural language without changing meaning\\
{\bf FSD} & Different ways of saying ``differentiate speech synthesis and voice conversion speech from real speech'' in 15 ways in a natural language without changing meaning\\ 
{\bf ER} & Different ways of saying ``emotion recognition of spoken utterance'' in 15 ways in a natural language without changing meaning\\  
{\bf AcC} &  Different ways of saying ``identifying the accent in spoken utterance'' in 15 ways in a natural language without changing meaning\\    
{\bf SD} & Different ways of saying ``detect whether given speech has sarcasm'' in 15 ways in a natural language without changing meaning\\ 
{\bf GID} &  Different ways of saying ``detect gender of speaker in given speech'' in 15 ways in a natural language without changing meaning\\ 
{\bf VAD} & Different ways of saying ``detect if there is speech in the given audio'' in 15 ways in a natural language without changing meaning\\ 
{\bf AuC} & Different ways of saying ``classify the environment sound present in audio'' in 15 ways in a natural language without changing meaning\\   
{\bf NER}&  Different ways of saying ``recognise the named entities in spoken utterance'' in 15 ways in a natural language without changing meaning\\   
{\bf SP} &  Different ways of saying ``write the semantic parse of spoken utterance'' in 15 ways in a natural language without changing meaning\\
{\bf DAC} &  Different ways of saying ``classify the dialog act of spoken utterance'' in 15 ways in a natural language without changing meaning\\
\bottomrule
\end{tabular}
}
\caption{Prompts for ChatGPT to generate task descriptions for each task type.}
\label{tab:chatgpt_prompt_detail}
\end{table*}
\begin{table*}[!htp]\centering
\scriptsize
\resizebox{\linewidth}{!}{%
\begin{tabular}{p{0.03\linewidth} | p{0.05\linewidth} | p{0.63\linewidth}}
\toprule
\textbf{Task Type} &\textbf{Split} &\textbf{Natural Language Phrase} \\\midrule
\multirow{17}{*}{\textbf{IC}} & \multirow{15}{*}{Train/Dev} & Assessing and categorizing the meaning behind spoken utterances \\
 & & Uncovering the intent behind verbal communication \\
 & & Sorting and classifying the purpose of spoken statements \\
 & & Decoding and categorizing the intention behind spoken language \\
 & & Understanding and labeling the intent in spoken expressions \\
 & & Organizing and categorizing the meaning of spoken utterances \\
 & & Interpreting and classifying the purpose of verbalized intentions \\
 & & Assigning labels to classify the intent of spoken communication \\
 & & Analyzing and categorizing the underlying purpose of spoken words \\
 & & Grouping and identifying the intention behind spoken utterances \\
 & & Differentiating and classifying the intent expressed through speech \\
 & & Extracting and labeling the classification of spoken intentions \\
 & & Segregating and classifying the purpose of verbalized expressions \\
 & & Unraveling and categorizing the intended meaning of spoken statements \\
 & & Intent classification of spoken utterance \\
 \cmidrule{2-3}
 & \multirow{5}{*}{Test} & Classifying the purpose of verbal expression  \\
 & & Identifying the intent conveyed through speech \\ 
 & & Decipher the objective conveyed through speech \\ 
 & & Determine the purpose of the spoken expression \\ 
 & & Identify the motive embedded in speech \\ 
 \midrule
 \multirow{17}{*}{\textbf{SCR}} & \multirow{15}{*}{Train/Dev} & Recognizing speech command \\
 & & Decoding oral requests \\
 & & Grasping speech guidance \\
 & & Discerning spoken triggers \\
 & & Acknowledging vocal commands \\
 & & Comprehending voiced instructions \\
 & & Interpreting verbalized prompts \\
 & & Detecting spoken signals \\
 & & Capturing voice-activated orders \\
 & & Classifying speech-based commands \\
 & & Inferring uttered guidance \\
 & & Analyzing vocal instructions \\
 & & Differentiating voiced prompts \\
 & & Sorting vocalized guidance \\
 & & Parsing spoken directives\\
  \cmidrule{2-3}
 & \multirow{5}{*}{Test} & Understanding spoken instructions \\
 & & Identifying uttered directives \\ 
 & & Identify spoken instructions \\ 
 & & Understand voiced directives \\
 & & Decode vocal orders \\
 \midrule
  \multirow{17}{*}{\textbf{ER}} & \multirow{15}{*}{Train/Dev} & Emotion recognition of spoken utterance \\
 & & Understanding the emotional content of spoken utterance \\
 & & Interpreting and categorizing emotions in verbal communication \\
 & & Decoding the emotional aspect of spoken expressions \\
 & & Discerning and recognizing emotions in spoken language \\
 & & Assessing and labeling the emotional state of spoken utterances \\
 & & Grasping the emotional intent behind spoken words \\
 & & Uncovering and categorizing the emotional response in speech \\
 & & Extracting and identifying the emotions conveyed through spoken utterances \\
 & & Emotionally analyzing and categorizing spoken expressions \\
 & & Capturing and recognizing emotional cues in spoken utterance \\
 & & Processing and discerning emotions in verbalized communication \\
 & & Differentiating and classifying the emotional tone of spoken utterances \\
 & & Unraveling and categorizing the intended emotional expression of spoken statements \\
 & & Inferring and organizing the classification of emotions in spoken language \\
  \cmidrule{2-3}
 & \multirow{5}{*}{Test} & Detecting emotional content in spoken utterance  \\
 & & Identifying emotions conveyed through speech \\
 & & Perceive the emotion conveyed through speech \\ 
 & & Discern the feelings embedded in spoken utterance \\ 
 & & Categorize the emotional content of verbal expression \\ 
\bottomrule
\end{tabular}
}
\caption{Detail of Natural Language Phrases used as Prompts to train and evaluate our MTL model in Table 2 }\label{tab:nlp_prompt_detail}
\end{table*}

\end{document}